\documentclass[conference]{IEEEtran}
\IEEEoverridecommandlockouts
\usepackage{cite}
\usepackage{amsmath,amssymb,amsfonts}
\usepackage{algorithmic}
\usepackage{graphicx}
\usepackage{amsthm}
\usepackage{censor}
\usepackage{hyperref}
\usepackage{textcomp}
\usepackage{tabularx}
\usepackage{titlesec}

\usepackage{xcolor}
\usepackage{float}
\usepackage{array}
\usepackage{amsmath}
\usepackage{graphicx}
\usepackage{booktabs}
\usepackage{subcaption}

\newtheorem{theorem}{Theorem}
\def\BibTeX{{\rm B\kern-.05em{\sc i\kern-.025em b}\kern-.08em
    T\kern-.1667em\lower.7ex\hbox{E}\kern-.125emX}}
\begin{document}

% \IEEEoverridecommandlockouts
% \IEEEpubid{\makebox[\columnwidth]{979-8-3503-0955-3/23/\$31.00~\copyright2023 IEEE \hfill} \hspace{\columnsep}\makebox[\columnwidth]{ }}

% Add the following code after the \maketitle command:
% \IEEEpubidadjcol 

\title{Addressing Large Action Spaces in 3D Floorplanning via Spatial Generalization  \\
% \thanks{* denotes equal contribution.}
}

%%%%%%%%%%%%%%%%%%%%%%%%% start real authors
\author{%
  Fin Amin,
  Nirjhor Rouf,
  Tse-Han Pan,
  Sounak Dutta,
  Md Kamal Ibn Shafi,
  Paul D. Franzon\\
  Department of Electrical and Computer Engineering\\
  North Carolina State University\\
  \texttt{\{samin2, nrouf2, tpan, sdutta6, mibnsha, paulf\}@ncsu.edu}
}

%%%%%%%%%%%%%% end real authors%%%%%%%%%%%%%%%%%%%%%

%%%%%%%% start anon authors
% \author{\IEEEauthorblockN{Anonymous Authors}}
% \and
% \IEEEauthorblockN{Anonymous Author 2}
% \IEEEauthorblockA{\textit{Department of Anonymous} \\
% \textit{Anonymous University} \\
% \textit{City, State, Country}\\
% email2@domain.com}
% \and
% \IEEEauthorblockN{Anonymous Author 3}
% \IEEEauthorblockA{\textit{Department of Anonymous} \\
% \textit{Anonymous University} \\
% \textit{City, State, Country}\\
% email3@domain.com}
% }
%%%%%%%%%%%%% end anon authors

\maketitle

\begin{abstract}
Many recent approaches to ML-based floorplanning represent the problem using discrete canvas coordinates, leading to scalability bottlenecks as action spaces grow. In this work, we explore the impacts of learning a \textit{continuous action representation} for 3D floorplanning tasks. By reasoning over the continuous placement space and discretizing only at inference time, our model decouples output shape from canvas resolution--enabling tractable learning and inference over large design spaces. A key insight behind our approach is the principle of \textit{$L$-action similarity}: the return of a given action is often similar to that of nearby actions in the placement space. This smoothness structure distills an inductive bias humans intuitively rely on when making physical placement decisions. By exploiting this structure, we improve how much our model learns about a decision based on similar decisions. As a case study, we show that this approach can learn how to construct floorplans even when pre-trained on random floorplans. Our results highlight the advantages of continuous decision spaces to address the issue of floorplanning's large action space.

\end{abstract}

\begin{IEEEkeywords}
floorplanning, continuous action space, decision transformer, spatial generalization, electronic design automation, reinforcement learning
\end{IEEEkeywords}

\section{Introduction and Motivation}
%I. Floorplanning is difficult and critical

Floorplanning is a major challenge in integrated circuit (IC) design that is pivotal in meeting the design specifications. The placement of major functional blocks can have a significant impact on the IC's power, performance, area, and thermal specifications. As a result, the quality of the floorplan is critical as poor floorplans are often expensive to fix once the subsequent physical design processes are initiated. Exacerbating these challenges is the fact that floorplanning is an NP-hard problem\cite{vlad} which will often exhibit a large search space for optimization even with a moderate number of functional blocks. These requirements on floorplan quality and scalability have motivated research on machine learning solutions to generate desirable floorplans.

\begin{figure}[ht]
  \centering
  \includegraphics[width=0.95\linewidth]{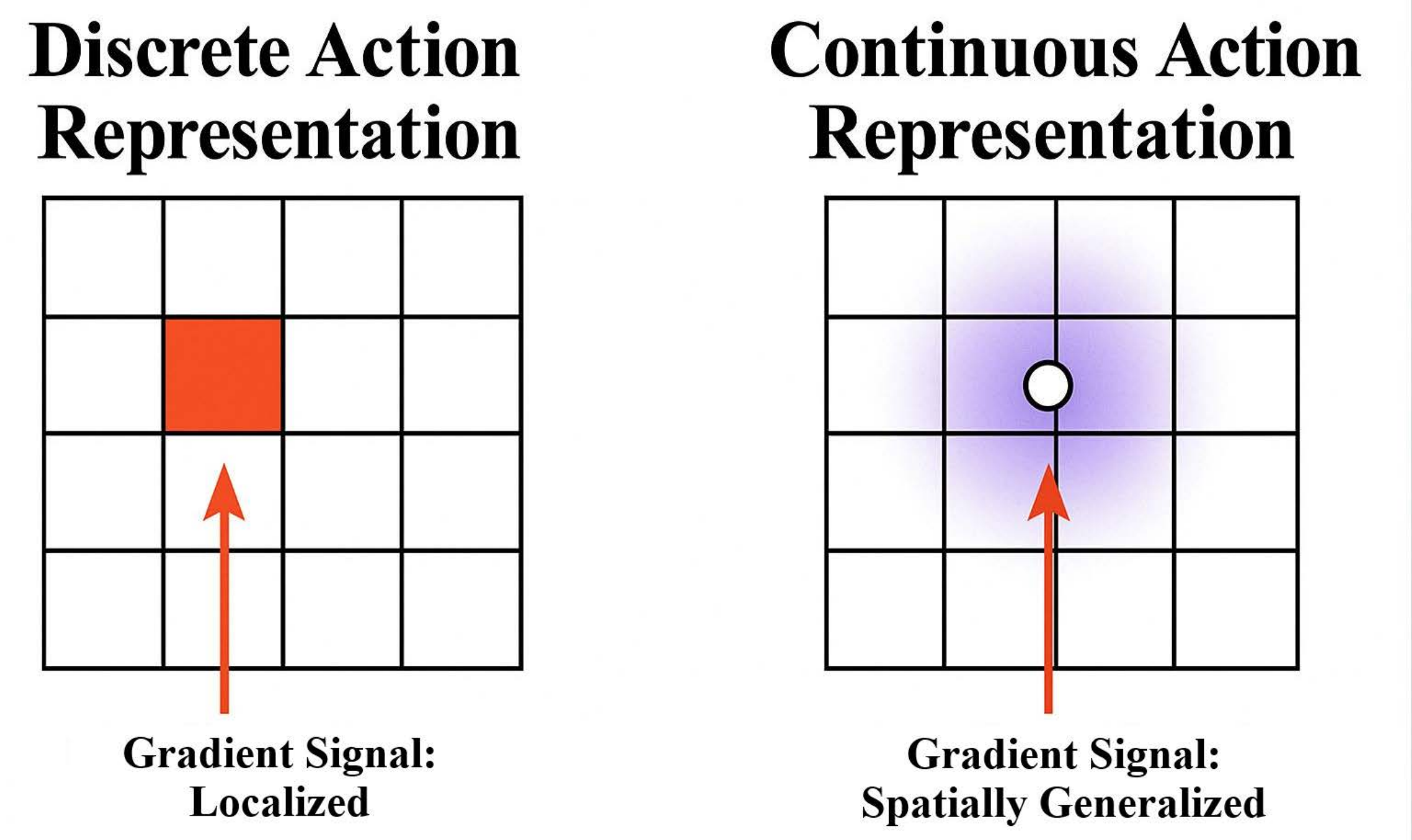}
  \caption{
A conceptual illustration of discrete vs. continuous action representations in floorplanning. In discrete settings (left), the placement model receives gradient feedback only for the selected grid location—limiting learning to that specific coordinate. In contrast, continuous representations (right) allow the model to generalize feedback across nearby spatial placements, enabling more efficient learning of physical layout patterns. This spatial inductive bias mirrors the intuition that human designers often rely on: that small changes in module position tend to produce similar design outcomes, such as wirelength or congestion.
  }
  \label{fig:gradient_feedback}
\end{figure}

%II. related works

Recent works have explored the use of reinforcement learning (RL) in placement, reporting improved placement times and competitive layout quality compared to human-designed baselines \cite{mirhoseini, cheng, chipformer}. These studies establish that RL is a viable approach for floorplanning. Notably, \textit{CircuitTraining}\cite{mirhoseini} demonstrated that deep RL can generate complete chip floorplans, motivating a series of follow-ups such as \textit{Chipformer} \cite{chipformer}, which casts chip placement as a sequence modeling problem. However, these approaches--while innovative--remain tightly bound to a discrete grid representation of the canvas. In Chipformer, the output action space is a classification over grid locations, requiring the model to scale linearly with canvas resolution. As remarked by prior work, this quantization limits both scalability and generalization \cite{wolpertinger}.

A central contribution of this paper is to recognize and formalize module placement as an instance of a structured large discrete action space (SLDAS), a setting where many actions exhibit local similarity in their return. In other domains, identifying decision spaces as an SLDAS has enabled more tractable analysis and algorithmic design~\cite{high_luminosity}.  To the best of our knowledge, this is the first time such a mathematical formalization has been introduced in the context of chip layout. \textbf{Our focus is therefore foundational: rather than proposing a complete floorplanning algorithm or benchmarking a suite of techniques, we examine how different action representations---continuous versus discrete---affect generalization.} 

Specifically, we ask whether a continuous representation can encode the same inductive bias that human designers intuitively exploit when reasoning about physical space, namely, that nearby placements tend to yield similar outcomes. We revisit this formally in Section II-C under the notion of $L$-action similarity, which constrains how the value of a decision varies across spatial locations. Our aim is not to optimize floorplans under ideal supervision, but to investigate whether this inductive structure is itself sufficient to support meaningful learning, even when trained on sub-optimal or randomly generated trajectories.

We refer to our technique as Spatial Generalization for Floorplanning (SGF), a method that explores continuous representations as a means of encoding inductive structure. This structure allows the model to generalize over spatial layouts without binding outputs to fixed grid locations. % Our work is inspired by recent trends in neural algorithmic reasoning and research on aligning human-robot decision making \cite{Neural_algorithmic_reasoning,generalist_neural_algorithmic_learner, what_can_nns_reason_about, aligning_robot_human_rep, getting_aligned_on_representational}.

To evaluate this setup, we train SGF entirely on random floorplan trajectories. This setup allows us to test whether the model can learn to infer meaningful placement behaviors--not from expert data, but from its own inductive bias about how similar placements yield similar outcomes. We connect this to the idea of $L$-action similarity, a structure also present in human reasoning over physical spaces.

%Moreover, this work not only uses interconnect wirelength, but also explores congestion and thermal estimations as optimization objectives. Namely, we show how SGF addresses multiple objectives when considering the return of a decision demonstrating its relevance to multi‑objective EDA challenges.

\section{Background}
Our discussion of prior work has two central themes. The first theme is an overview of existing ML-based approaches to floorplanning. The goal is to discuss their contributions along with their limitations. The second theme concerns the relevant discoveries in the field of RL and auto-regressive models which guided our algorithm design.

\subsection{ML Approaches to Floorplanning}

The classical approaches for 2D floorplanning usually leverage analytical and metaheuristic algorithms. While analytical approaches are good at finding the minimum area solutions, they scale poorly with a large number of constraints. On the other hand, metaheuristic methods are capable of handling more complex problems, however, the lack of knowledge re-use makes them quite inefficient.

The issues with the classical algorithms motivated the search for more efficient solutions; one pioneering step towards this was \cite{mirhoseini}. This work showed the feasibility of approaching the floorplanning problem from the perspective of RL. Their off-the-shelf RL algorithms while successful in implementation are not well-optimized. Their agent learns slowly despite high computational resources.

Other RL-based approaches to floorplanning have explored alternative problem representations, such as sequence pairs and corner block lists \cite{goodfloorplan, corner_block_list_FP_RL}. While these representations are effective for certain formulations of 2D floorplanning, they do not readily extend to 3D floorplanning, which introduces new spatial and structural challenges hence they are not the focus of our work.

\textit{Chipformer},\cite{chipformer} shows promising results in learning transferable policy placement through offline training and online fine-tuning of a decision transformer. This method improves efficiency compared to previously published works however, they do not leverage learning the spatial structure of the problem itself. Moreover, while Chipformer achieves strong results on wirelength, it relies on expert trajectories for offline pretraining. Similar to the aforementioned, \textit{FlexPlanner} \cite{flexplanner} introduces a reinforcement learning framework that discards heuristic search and directly operates in a hybrid action space. Their policy network outputs a probability map over canvas positions, a distribution over die layers, and parameters for block aspect ratios, enabling the model to explicitly account for 3D alignment and soft-block shaping. While this design achieves strong improvements on alignment and wirelength metrics, its reliance on discrete position maps ties output dimensionality to canvas resolution, hindering scalability.

\subsection{RL and Decision Transformers}

 In RL, the environment represents the problem or the space in which the agent operates. It manages the state, action, and reward transitions of the problem the agent is trying to solve. The agent improves its policy through its interaction with the states and rewards returned by the environment. More specifically, at each timestep, the agent selects an action (such as where to place a module) and the environment returns a reward (such as wirelength) for the action chosen and the next state. The objective is to learn a policy, $\pi^*(a|s)$, that learns to select an action, $a$, in the action space, $\mathcal{A}$, given a state, $s$, which maximizes the cumulative reward over time. 

The action-value function, otherwise known as the Q-function, $Q^\pi(s, a)$ under policy $\pi$, is the expected return after taking an action $a$ in state $s$ and then following policy $\pi$ \cite{sutton_rl}. Typically, this Q-function is parameterized by some estimator (typically a neural network), giving us $Q_\theta^\pi(s, a)$. Similarly, the policy $\pi$ is also parameterized by some estimator, $\pi_\theta$. The agent interacts with the environment using its policy, and the Q-function guides the agent by estimating the value of actions.

In actor-critic algorithms, this setup is formalized by having two distinct components: the actor and the critic. The actor selects actions using a learned policy $\pi_\theta(a \mid s)$, while the critic estimates the action-value function $Q_\theta^\pi(s, a)$ to guide policy improvement. The critic provides feedback on the actor’s choices by approximating $Q^\pi(s, a)$, typically using temporal-difference learning \cite{actorCriticOG, sutton_rl}.

Decision transformers (DTs) \cite{decisionTransformer} extend transformers \cite{attentionisAllYouNeed} to decision-making problems. The idea is to learn offline from \textit{trajectories}, 
\[
\Gamma= \{(s_0, a_0, r_0), (s_1, a_1, r_2), \ldots, (s_T, a_T, r_{T})\},
\]
and the \textit{rewards-to-go} (RTGs), where
\[
g_t = \sum_{m=t}^{T} r_{m+1}.
\]
In other words, $g_t$ is the cumulative future rewards that the agent aims to achieve from a given timestep until the end of the trajectory. We use the subscript $\gamma_t$ to denote the \textit{trajectory-so-far}, which includes the states, actions, and rewards/rewards-to-go (SARs) seen up until $t$. This corresponds with $\Gamma[0:t-1]$\footnote{We slightly abuse notation here. For the actor, we time-delay all the actions by one step. In other words, the actor would see all actions up until $a_{t-1}$, but would see $s_t$ and $g_t$. We forgo this time-shifting for the critic, as the critic is used to predict the return-to-go for $\hat a = \pi(s_t, a_t, g_t).$}.

The idea is to ``prompt'' the model into producing the desired actions in an auto-regressive manner. At the first timestep, the model is given a target cumulative reward value, which is the reward we aim to achieve after all decisions have been made. For each subsequent timestep: $\hat{a} = \mathtt{DT}(\gamma_t, g_t)$, and the loss is computed between $\hat{a}$ and $a_t$. Note that $g_t$ is computed by subtracting the rewards the agent has seen so far from the target reward. Online DT training works similarly. Trajectories are collected on policy in a replay buffer, and the likelihood of their occurrence is maximized using a max-entropy sequence modeling loss. For brevity, we refer readers to \cite{onlineDecisionTransformer} for more details.

\subsection{Reasoning in Structured Large Discrete Action Spaces}

The floorplanning task involves placing various functional blocks on an integrated circuit to optimize multiple design criteria such as wirelength, congestion, and thermal performance. Each block can occupy numerous potential positions, leading to a combinatorial explosion of possible configurations. This interdependence and the vast number of possible placements make traditional RL methods inefficient for such problems. To make things worse, existing approaches are constrained by how they discretize the floorplan canvas.

For example, a canvas discretized to $100 \times 100$ locations corresponds to $|\mathcal{A}| = 10^4$ possible placement coordinates. A model operating in this discrete action space must produce a probability distribution over all possible actions (a \textit{softmax} over the canvas). This requires $10,000$ output units just to compute the logits for each location during inference. Extending this to 3D with just two vertical layers increases the action space to $|\mathcal{A}| = 2\times10^4$, necessitating a further $2\times$increase in the number of output units. The size of these action spaces can make ML models too large to run on many hardware configurations. Compute concerns aside, the extreme size of the action space can make ordinary RL approaches not very sample efficient \cite{learned_action_representations, wolpertinger}. 

To ameliorate this issue, we observe that floorplanning operates on a structured large discrete action space (SLDAS). SLDAS is a mathematical formalism introduced in RL literature to characterize large, spatially organized decision spaces. The salient feature of these spaces is that they can exhibit a form of structure known as $L$-\textit{action similarity} \cite{dynamicneighborhood}, where $L$ quantifies how much the action-value function $Q^{\pi}(s, a)$ can change in response to changes in the action $a$. Formally, $L$ is defined as:

\begin{equation*}
L = \sup_{\substack{a,a' \in A' \\ a \neq a'}} 
\frac{
\textcolor{purple}{|Q^{\pi}(s, a) - Q^{\pi}(s, a')|}
}{
\textcolor{blue}{\|a - a'\|_2}
}
\end{equation*}

ensuring that

\[
|Q^{\pi}(s, a) - Q^{\pi}(s, a')| \leq L \|a - a'\|_2 \quad \text{for all } a, a' \in A'.
\]

This constraint implies that changes in the action space result in bounded changes in expected return. In other words, the \textcolor{purple}{difference in reward between two potential locations to place a module} is bounded proportionally to the \textcolor{blue}{difference between the locations themselves}. We conjecture that this is an important inductive bias to learn for the floorplanning task.

As discussed, standard RL models struggle to scale in such structured large discrete action spaces. In particular, outputting a probability distribution over all possible actions becomes computationally expensive, and models fail to generalize across nearby actions due to the lack of spatial inductive bias.

The \textit{Wolpertinger} agent \cite{wolpertinger} addresses this by exploiting action similarity. By combining continuous action embeddings with $k$-nearest neighbor search, the agent is able to operate in a continuous domain while ensuring that selected actions are valid discrete placements. The benefit of this design choice is that the output shape of the model is no longer tied to the size of the action space. For example, in the same  3D layout setting with a $2 \times 100 \times 100$ canvas, a discrete model would require $20{,}000$ output logits to represent the full action space. In contrast, the Wolpertinger agent outputs a point in $\mathbb{R}^3$, normalized to $[0,1]^3$, corresponding to $(x, y, z)$ coordinates. This decouples the output size from canvas resolution while retaining spatial expressiveness.

% \subsection{Addressing Over-estimation Bias}

\section{Spatial Generalization for Floorplanning}

In this section, we describe architecture behind SGF by discussing the relevant prior work that guided our design process. A key observation we make is that placing a module in location $x, y, z$ is likely to give a similar return as placing a module in location $x+\epsilon, y+\epsilon, z+\epsilon$. This effect is a direct consequence of $L$-action similarity—nearby actions are expected to produce similar returns (see Section II-C). 

Figure~\ref{fig:gradient_feedback} illustrates a central advantage of continuous action representations: they allow gradient signals to propagate not only to the chosen action, but also to nearby alternatives. This facilitates generalization through spatial smoothness. In discrete classification setups, such as those used in Chipformer or CircuitTrainer, the model receives reward feedback only for the chosen action--there is no gradient signal for neighboring grid locations. This creates sparsity in reward learning in large action spaces. In contrast, continuous action representations allow the model to receive a gradient signal for the surrounding region in the action space. This inductive bias enables smoother learning and generalization from noisy or randomly generated data, as similar actions tend to yield similar outcomes in spatial domains like floorplanning.

\begin{figure*}[th] 
  \centering
  \includegraphics[trim=0cm 1cm 0cm 0cm, clip=true,width=.75\linewidth]{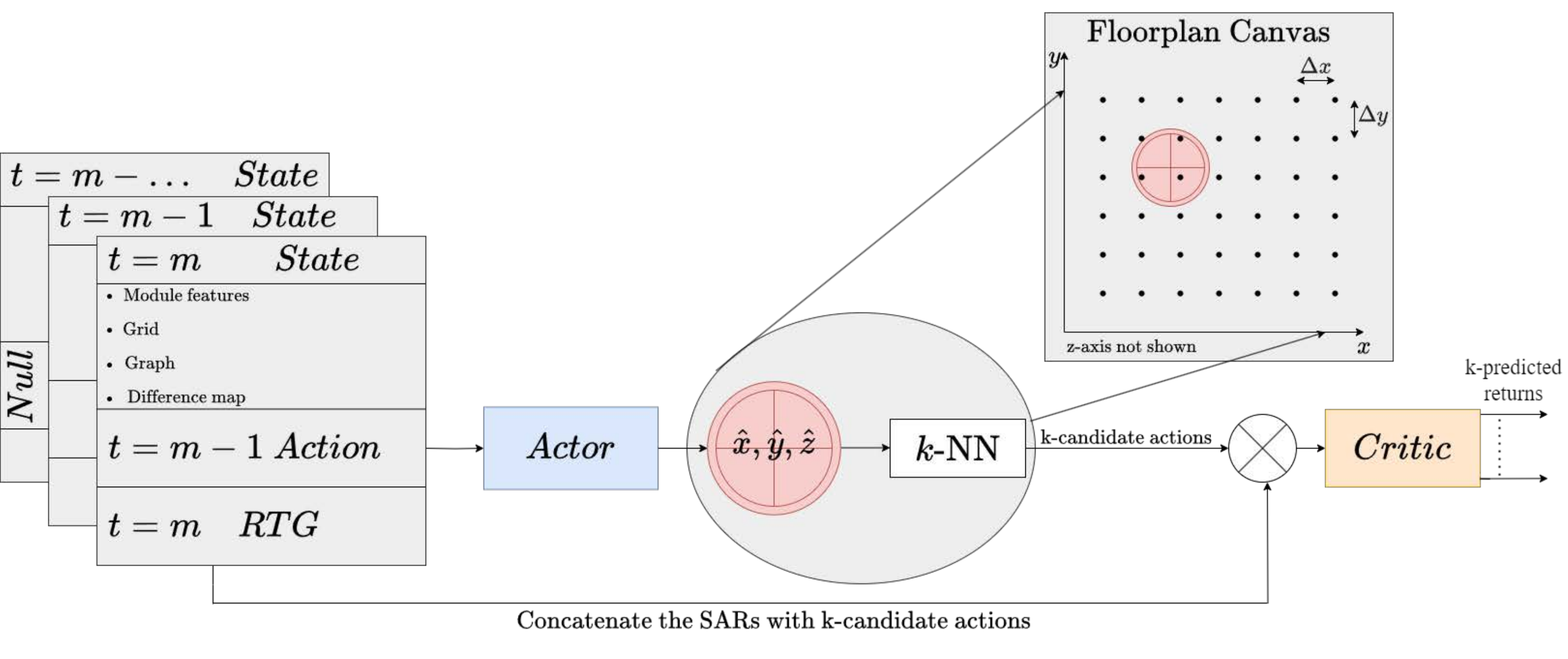}
  \caption{Our agent melds decision transformers with the Wolpertinger architecture. This figure shows the flow of information across a single time-step of inferencing. First, the SARs ($\gamma_t$) are input into the actor so that the actor can propose a candidate centroid location $\hat{\alpha}_t$, which is mapped via $k$-NN to $k$ nearby legal discrete actions ${\hat{a}_t^{(1)}, \ldots, \hat{a}_t^{(k)}}$. The critic then evaluates each and selects the final action $\tilde{a}_t$ by minimizing return-to-go prediction error (eq. \ref{eq:critic_argmax}).}
  \label{fig:LRM_architecture.png}
\end{figure*}

\subsection{SGF Architecture}

SGF follows the general structure of the Wolpertinger agent, which combines continuous action prediction with $k$-nearest neighbor search to enable decision-making in large discrete action spaces. In our adaptation, we replace the traditional neural networks used in the Wolpertinger with decision transformers as both the actor and critic modules. The model is composed of three components: a DT-based actor $\pi_\theta$, a DT-based critic $g_\theta^\pi$, and a $k$-NN module \cite{knn} for action selection. We first describe the information flow between these components, followed by the input embeddings used to support the model.

At each timestep, the actor outputs a point $\hat\alpha_t \in [0,1]^3$ representing the predicted normalized centroid location for the next module placement. The $k$-NN module then maps this to the nearest $k$ legal discrete actions $\{\hat a_t^1, \dots, \hat a_t^k\} \subset \mathcal{A}$, where $a_t^i \in \mathbb{Z}^3$. Here, $\mathcal{A} \subset \mathbb{Z}^3$ is the set of all valid placement coordinates on the discrete floorplanning grid.

We denote the $k$-NN neighborhood by
\[
\mathcal{N}_k(\hat{\alpha}_t) \triangleq \{\hat a_t^{(1)},\ldots,\hat a_t^{(k)}\}\subset\mathcal{A},
\]
and the nearest-proposal radius by
\[
\psi_k \triangleq \min_{a\in \mathcal{N}_k(\hat{\alpha}_t)} \|a-\hat{\alpha}_t\|_2 .
\]
Afterward, these $k$ actions are fed into the critic. Figure \ref{fig:LRM_architecture.png} shows this flow of information. The critic predicts the return-to-go of the $k$ candidate actions. The action which most closely achieves the target $g_t$ is ultimately selected. That is, we select:

\begin{equation}
\tilde{a}_t = \arg \min_{a \in \{\hat{a}_t^{(1)}, \ldots, \hat{a}_t^{(k)}\}} \left| g_\theta^\pi(\gamma_t, a) - g_t \right|_1
 \label{eq:critic_argmax}
\end{equation}
The output shape of both the actor and critic is $3$. However, their activation functions differ. For the actor, we scale all $\alpha = \hat{x},\hat{y},\hat{z}$ coordinates between $0.0$ and $1.0$ via $0.5 \times (\tanh(\texttt{actor\_logits}) + 1)$. In our case, due to all returns-to-go being positive, the critic's logits are activated via $\mathrm{ReLU}(\texttt{critic\_logits})$.

\begin{figure}[ht] 
  \centering
  \includegraphics[trim=0cm 0cm 0cm 0.3cm, clip=true, width=0.9\linewidth]{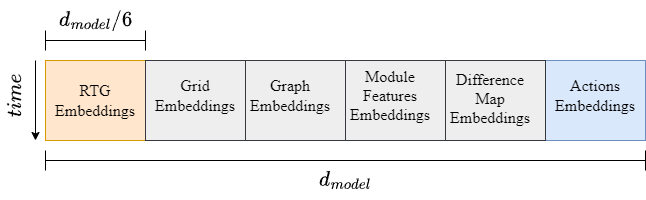}
  \caption{The shape and positions of the input embeddings as they are concatenated in the $\mathrm{token}$ $\mathrm{embeddings}$. The gray components are composed of the state. We also add a time-wise positional embedding to the entire token (not shown). }

  %Note that we differ from existing work by considering this entire $d_{model}$-sized token embedding for a single timestep. Other work typically discretizes \textit{states}, \textit{actions}, \textit{rewards} into sub-timesteps.}

  %It should be noted that the actor sees all the past actions taken up to the current timestep whereas the critic sees all the actions including the current timestep
 
  \label{fig:embedding}
  
\end{figure}

The three key inputs to the actor and critic are \textit{RTGs}, \textit{states}, and \textit{actions}. Table \ref{table:input_encoding} shows the encoding methods for the different inputs and Figure \ref{fig:embedding} shows how they are embedded as tokens for both decision transformers. There are four components to the state embedding which are generated at each time step:

\begin{enumerate} \label{list:states}
    \item $\mathtt{Grid.}$ This contains two components: the locations of the already placed modules (the canvas) and--inspired by \cite{chipformer}--the wirelength increase for placing a module at a particular location. 
    \item $\mathtt{Graph.}$ Netlists are processed by a graph attention network~\cite{Graphattentionnetwork}. At each timestep, we embed the difference between the full netlist graph and the subgraph placed so far on the canvas, using area, height, width and degree as node attributes:
    \begin{equation}
        \mathcal{G}_{embed}(t) = \mathtt{GAT}(\mathcal{G}_{netlist}(t)) - \mathtt{GAT}(\mathcal{G}_{placed}(t)).
    \end{equation}

    \item $\mathtt{Difference Map.}$ The difference map measures the change in the floorplan canvas across two consecutive timesteps. 
\end{enumerate}

% The RTG encoder uses scaled dot product attention \cite{attentionisAllYouNeed} to embed $g_t$; we discuss this decision in more detail in Table \ref{table:input_encoding}. 

Note that for the critic, we append the $k$ candidate actions returned by $k$-NN across the batch-axis and duplicate all the state and RTG information. This allows us to compute equation \ref{eq:critic_argmax} to select the current timestep's action. All the embedding mechanisms have an output shape of $\frac{1}{6}\cdot d_{\text{model}}$ as depicted in Figure \ref{fig:embedding}. We set the context length to $\mathrm{n\_modules} + 1$; we use the same parameters for both the actor and critic. Our $d_{\text{model}} = 192$, with $4$ attention layers and $4$ heads. Additional information can be found in the Appendix.

\subsection{What Can We Learn From Random Floorplans?}

Typically, in order to perform offline training, we would usually need expert trajectories $\Gamma^* \mathtt{ \sim } \pi^*$ which emulates the decision-making of an optimal policy. This is problematic because this necessitates either human experts to construct a dataset of trajectories or we must use an existing floorplanning tool which we would treat as an ``expert.'' Chipformer takes the latter approach by harnessing the outputs of MaskPlace \cite{maskPlace} to create $\hat{\Gamma}^*$. 

For various reasons, obtaining an expert trajectory dataset can be infeasible or intractable. Especially in the case of floorplanning, verifying that a given floorplan trajectory is optimal cannot be done efficiently unless P = NP. This is a direct consequence of the floorplanning being NP-hard. To investigate what we can learn from spatial generalization, we forgo collecting an expert trajectory dataset and \textbf{train on entirely random trajectories}, $\Gamma_{\xi}$. We obtain $\Gamma_{\xi}$ by randomly selecting a legal action from our floorplanning environment at each timestep and storing the resulting states and rewards. The goal is we want the model to learn patterns in how placements influence design objectives.

In our experiments, our actor is trained to match the actions of $\gamma$:

\begin{equation} \mathcal{L}_{\text{Actor}} = |\hat{\alpha}_t - \alpha_t^{\gamma}|_1 \label{eq:actor_loss} \end{equation}
where $\hat\alpha_t = \pi_\theta(\gamma_t)$ and $\alpha_t^{\gamma} \in [0,1]^3$ is the normalized continuous form of the ground-truth action $a_t$ from trajectory $\Gamma$. Similarly, the critic is trained to minimize the error in predicting the return-to-go:

\begin{equation} 
\mathcal{L}_{\text{Critic}} = |g_t^{\gamma} - g_\theta(\gamma_t)|_1 \label{eq:critic_loss} \end{equation}
The rationale behind training on random trajectories is given by hindsight experience replay (HER) and goal-conditioned supervised learning (GCSL)\cite{hindsight_experience_replay, learningtoReachGoals}. The idea behind HER and GCSL is to consider failed attempts as valuable learning experiences by relabeling the trajectories to be optimal in hindsight as though the data came from an optimal agent with a different goal. Less formally, we observe that random floorplans can be evaluated and used to train a model (the critic) to learn if a decision (placement of a module given the current floorplan and remaining modules) aligns with the design objectives.  

However, the loss for GCSL is for categorical action spaces: 

\begin{equation}
\mathcal{L}_{\text{GCSL}}(\pi) = -\mathbb{E} \left[ \sum_{t=0}^{T} \log \pi(a = a_t | \gamma_{t}) \right].
\end{equation}
This loss computes the maximum-likelihood estimate of $a_t$ given $\gamma_t$. In order to work in a continuous action space, we need to modify this objective. If we assume that the errors follow a Laplace distribution, the maximum likelihood estimate becomes $\mathcal{L}_{\text{Actor}}$ (eq. \ref{eq:actor_loss}) \cite{Detection_estimation_vantrees_part1}. An analogous mapping can be done between $\mathcal{L}_{\text{Critic}}$ (eq. \ref{eq:critic_loss}) and HER's training objective. Succinctly, the training procedure of our algorithm can be thought of as an adaptation of GCSL to train the actor and HER being used to train the critic.

Diverse random trajectories enable broad exposure to different layout configurations, allowing the model to observe how placements influence trade-offs across objectives such as wirelength, thermals, and congestion. While these trajectories are not optimized, they encode sufficient variability for the model to learn how design choices affect downstream metrics \cite{congestion_driven, wire_congestion}.

Another practical reason is the computational feasibility; generating random trajectories is straightforward and does not require the costly process of acquiring expert knowledge. Furthermore, this approach allows users to leverage existing drafts of floorplans that have already been evaluated for metrics. In other words, we permit normally ``sub-optimal'' samples to help us. Other work in RL has also remarked on the importance of sub-optimal trajectories. For example, the authors of \textit{Multi-Game DTs} have found that incorporating non-expert trajectories into training improved performance\cite{multi_game_DT}.

\subsection{SLDAS-Based Error Bound for SGF}
\label{subsec:sldas_bound}
We now derive a bound on SGF’s one-step suboptimality that explicitly exploits $L$-action similarity on an SLDAS.

\begin{theorem}\label{thm:sldas_bound_tight} \textbf{Error Bound under $L$-Action Similarity}

Assume $Q^\pi(s_t,\cdot)$ is $L$-Lipschitz on $A$:
\[
\textcolor{purple}{\big|Q^\pi(s_t,a)-Q^\pi(s_t,a')\big|} \;\le\;
L\,\textcolor{blue}{\|a-a'\|_2},\quad \forall a,a'\in A,
\]
and $\{a_t^\star\}\cup \mathcal{N}_k(\hat{\alpha}_t)\subseteq A$.
Let
\begin{align}
\textcolor{blue}{d_t} &\,\triangleq\, \min_{a\in \mathcal{N}_k(\hat{\alpha}_t)} \|a - a_t^\star\|_2,
\end{align}
and suppose the critic’s absolute RTG prediction error on $\mathcal{N}_k(\hat{\alpha}_t)$ is at most $\varepsilon_c$.
If $\tilde a_t\in\mathcal{N}_k(\hat{\alpha}_t)$ is the action selected by the critic, then
\begin{align}    
\underbrace{\textcolor{purple}{Q^\pi(s_t,a_t^\star)-Q^\pi(s_t,\tilde a_t)}}_{\Delta_t}
\;\le\;
L\,\textcolor{blue}{d_t}\;+\;2\,\varepsilon_c.
\end{align}
Moreover, since
\[
\textcolor{blue}{d_t}\;\le\;\textcolor{blue}{\|\hat{\alpha}_t-a_t^\star\|_2}
\;+\;\min_{a\in\mathcal{N}_k(\hat{\alpha}_t)}\textcolor{blue}{\|a-\hat{\alpha}_t\|_2},
\]
we recover the upper bound
\begin{align}
\Delta_t &\;\le\; L\,\big(\textcolor{blue}{\delta_t+\psi_k}\big) + 2\,\varepsilon_c, \label{eq:critic_hurts_q}\\
\textcolor{blue}{\delta_t} &\,\triangleq\,\|\hat{\alpha}_t-a_t^\star\|_2.
\end{align}

\end{theorem}

The SLDAS assumption links the \textcolor{purple}{return difference}
$\textcolor{purple}{|Q^\pi(s_t,a)-Q^\pi(s_t,a')|}$
to the \textcolor{blue}{action-space distance}
$\textcolor{blue}{\|a-a'\|_2}$ via $L$.
Therefore the optimality gap
\[
\Delta_t \;=\; \textcolor{purple}{Q^\pi(s_t,a_t^\star)-Q^\pi(s_t,\tilde a_t)}
\]
is at most $L$ times the \textcolor{blue}{distance} from the optimal discrete action to SGF’s $k$-NN candidate set
($\textcolor{blue}{d_t}$), plus $2\varepsilon_c$ for critic selection noise.
Decomposing $\textcolor{blue}{d_t}$ yields two intuitive pieces:
(i) \textcolor{blue}{$\delta_t$}--how far the actor’s continuous proposal $\hat{\alpha}_t$ is from the optimal discrete $a_t^\star$; and (ii) \textcolor{blue}{$\psi_k$}--how tightly the $k$-NN candidates cover the neighborhood of $\hat{\alpha}_t$. Thus smoother SLDAS structure (smaller $L$), better actor accuracy (smaller \textcolor{blue}{$\delta_t$}), denser/legal proposals (smaller \textcolor{blue}{$\psi_k$}), and a better-trained critic (smaller $\varepsilon_c$) all tighten the bound.

% \footnote{Why $\frac{\sqrt{3}}{2}h$? With an axis-aligned 3D grid of spacing $h$, space is partitioned into cubes of side length $h$ whose vertices are legal discrete actions. Any continuous point $\hat{\alpha}_t$ lies inside some cube; its nearest legal action is one of that cube’s $8$ vertices. The \emph{maximum} Euclidean distance to the nearest vertex occurs at the cube center, yielding
% $\sqrt{(h/2)^2+(h/2)^2+(h/2)^2}=\frac{\sqrt{3}}{2}h$.
% Thus $\textcolor{blue}{r_k}\le \frac{\sqrt{3}}{2}h$. 
% If spacings differ per axis ($h_x,h_y,h_z$), the bound becomes $\tfrac{1}{2}\sqrt{h_x^2+h_y^2+h_z^2}$; under $\ell_1$ (Manhattan) distance it is $\tfrac{3}{2}h$.}

\begin{figure}[h] 
  \centering
  \includegraphics[trim=0cm 0cm 0cm 0cm, clip=true,width=.825\linewidth, height=0.825\linewidth]{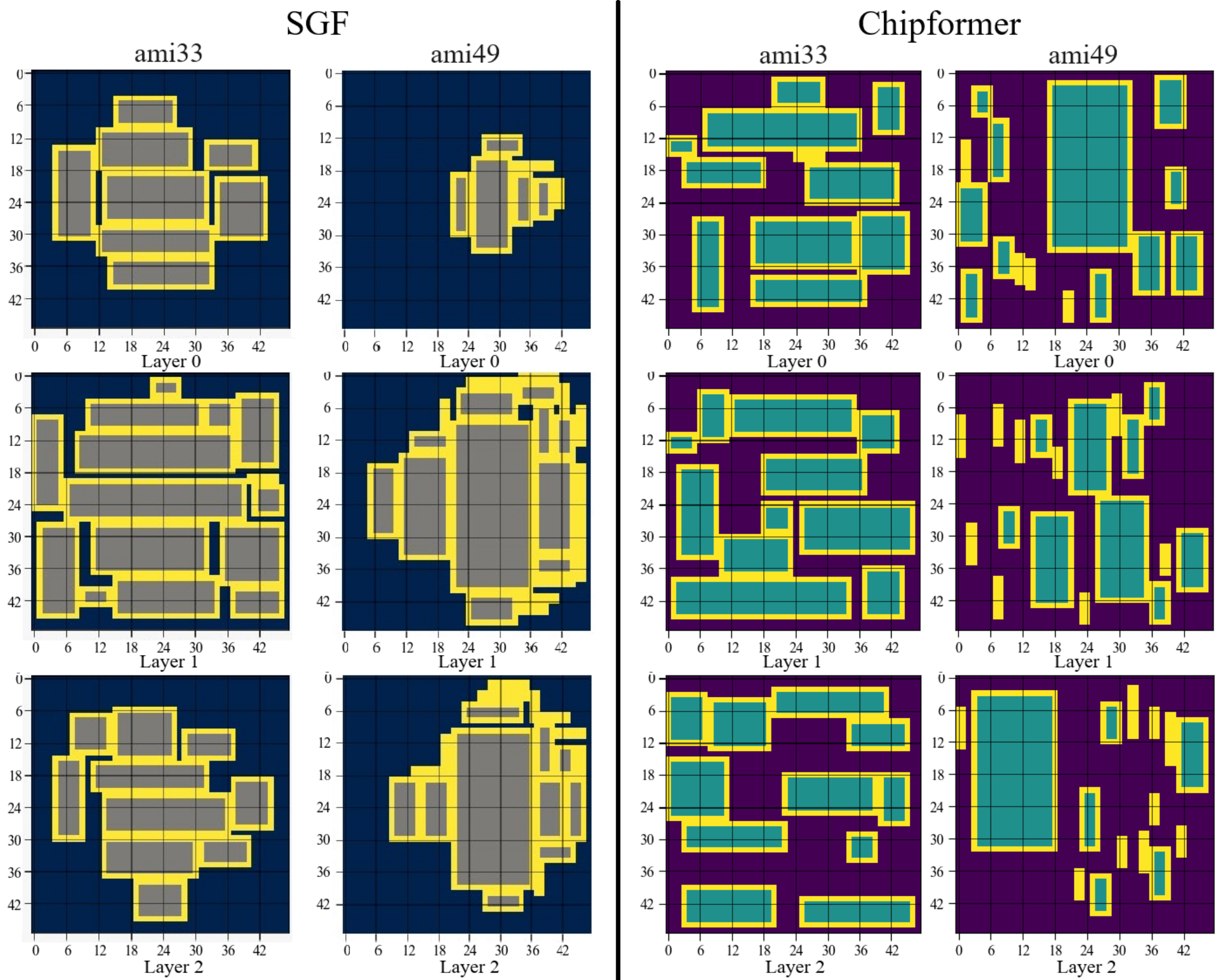}
  \caption{
Three-layer floorplans generated from the \textit{ami33} and \textit{ami49} netlists. SGF (left) produces compact and layered layouts with strong spatial regularity, despite being trained only on random trajectories. Quantitative performance metrics can be found in Table \ref{table:wirelength_results}.}

%The total Manhattan wirelength ($\downarrow$) of SGF is 200,738 and 28,636, compared to 225,877 and 39,442 for Chipformer, for \textit{ami33} and \textit{ami49}, respectively. 37184.0

  % \caption{
  % The following shows 3-layer floorplans generated from the \textit{ami33} and \textit{ami49} netlists. The total Manhattan wirelength ($\downarrow$) of SGF is 200,738 and 28,636 as opposed to 225,877 and 39,442 of Chipformer for the aforementioned netlists, respectively. The lower wirelength for \textit{ami49} can be explained by the latter having significantly fewer connections compared to \textit{ami33}. Note that Chipformer was trained offline and fine-tuned online whereas SGF was trained offline only.}
  \label{fig:all_FPs}
\end{figure}

\section{Experiments}

Rather than benchmarking against a broad set of algorithms, our goal is to explore how output representation—continuous versus discrete—affects generalization without expert trajectories. In this section, we examine how SGF behaves in comparison to Chipformer, a recent representative of discrete grid-based placement methods, as seen in earlier techniques such as CircuitTraining and FlexPlanner.

For our experiments, we produce an RL environment that works with the MCNC, IBM and GSRC benchmarks \cite{MCNC,ibm_fp_benchmarks, GSRC}. To do this, we configure our environment to return the states listed in \ref{list:states}. For parity with existing techniques, our environment returns wirelength--which has been the predominant performance metric for contemporary ML-based work. 

However, modern floorplanning often involves multiple objectives such as thermals, congestion, and wirelength. For the rewards, we address the multi-objective nature of floorplanning by returning three values at each timestep. $r_t = (w_t, c_t, h_t$) which correspond with a wirelength score ($\uparrow$), congestion penalty ($\downarrow$) and thermals penalty ($\downarrow$). The wirelength score is a proxy for the 3D Manhattan wirelength; it increases inversely with wirelength. These metrics allow us to evaluate the capabilities of SGF's critic component. To promote reproducibility, \textbf{we commit to releasing code after acceptance. }

\begin{figure}[t]
\centerline{\includegraphics[scale = .45]{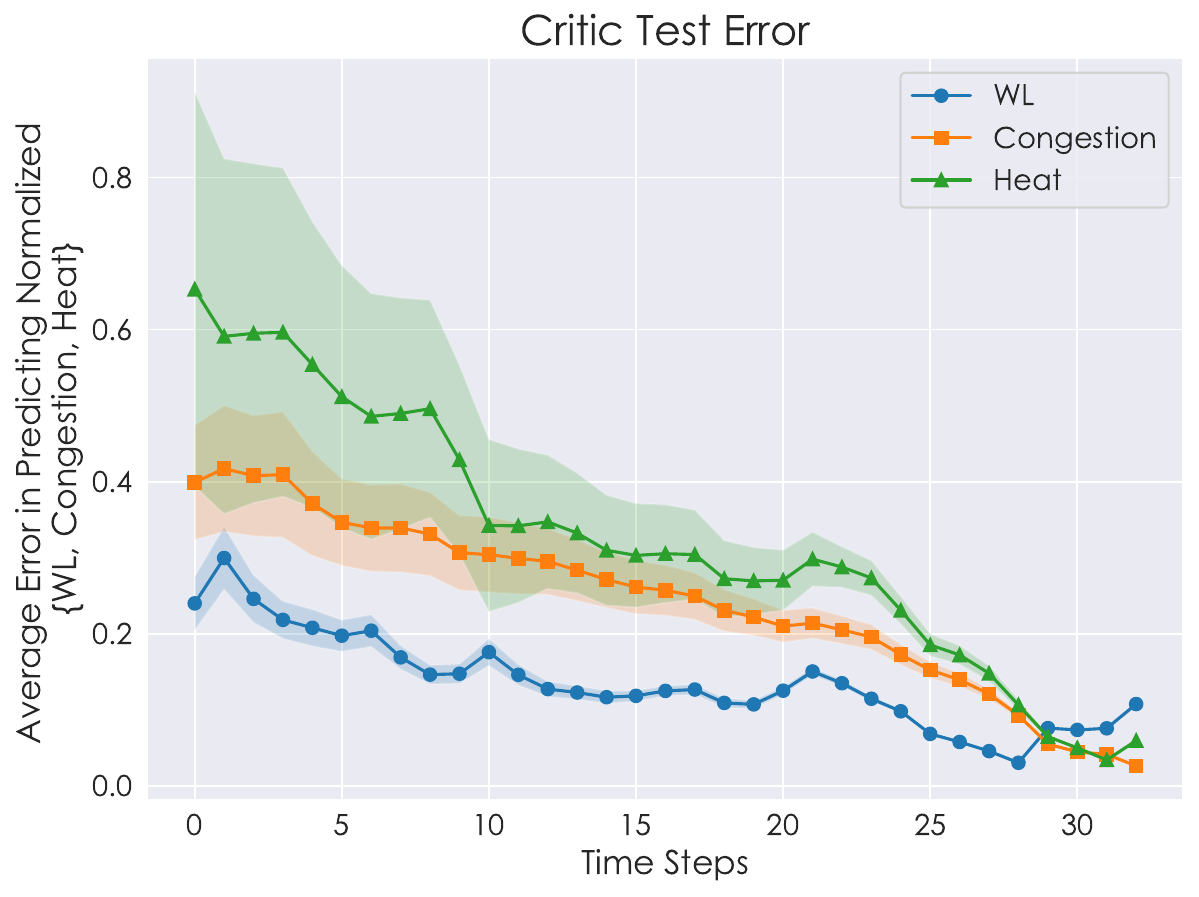}}
\caption{At each time step, a module is placed on the canvas and our environment returns three proxy metrics for wirelength (WL), congestion and thermals. This chart shows the SGF critic's MSE in estimating the ultimate performance of the design (the RTG). Shaded regions denote the variance.}
\label{fig:critic_test}
\end{figure}

\subsection{Evaluations}

As stated earlier, modern floorplanning often involves multiple objectives such as thermals, congestion, and wirelength. However, since Chipformer primarily evaluates wirelength, we align our experiments accordingly and use wirelength as the main metric for comparison.

Both methods are tested on the \textit{ami33, ami49, n50, n100} and \textit{ibm6} benchmarks. We prepared $|\Gamma_{\xi}| = 200$ offline trajectories for all of the netlists. Chipformer was additionally given a warm-up buffer of $|\Gamma_{\xi}| = 60$ trajectories to initialize its online fine-tuning phase. We define SGF’s return prompt using statistics from its training distribution: for each netlist, let $\mu_\gamma = (\mu_w, \mu_c, \mu_h)$ and $\sigma_\gamma = (\sigma_w, \sigma_c, \sigma_h)$ represent the mean and standard deviation of the returns in $\Gamma_{\xi}$. We then prompt SGF with $(\mu_w + 3\sigma_w, \mu_c, \mu_h)$ to guide trajectory generation toward higher wirelength scores while maintaining neutral targets for other metrics.

Our first experiment has two components. In the first component, Chipformer is trained both offline and online. Its online training is performed over 5 iterations, each consisting of 10 rollouts followed by fine-tuning. In contrast, SGF is evaluated purely after offline training with no online adaptation. This asymmetry is intentional: the goal is to assess whether structural generalization can emerge from a continuous action representation even without online training.

In the second component (conducted only on \textit{ami33}), we extend Chipformer’s online training to further probe the limits of discrete action representations. Starting from the model trained as described above, we perform 5 additional cycles of online fine-tuning. Each cycle consists of 5 iterations, and each iteration performs 10 rollouts followed by training. After each cycle—corresponding to 50 new online rollouts—we record and report the wirelength performance. This extended setup allows us to examine whether prolonged online learning allows Chipformer to bridge the generalization gap relative to SGF.

We perform a second experiment to assess whether the SGF critic captures structured reward trends as a trajectory unfolds. Specifically, we test whether it can consistently estimate return-to-go values based on partial observations of a floorplan. To do this, we trained the critic as in the wirelength experiment and then evaluated it on 30 unseen trajectories from \textit{ami33}. At each timestep, the critic outputs a predicted return-to-go given the trajectory-so-far, $\gamma_t$.

\section{Results and Discussion}

SGF achieves competitive wirelength results across all tested benchmarks, despite constraining its training (Table ~\ref{table:wirelength_results}). These results suggest that smooth spatial reasoning can provide a useful inductive structure--especially in settings where collecting high-quality training data is costly or infeasible.

% %\begin{table}[h]
% %\centering
% %\caption{Manhattan wirelength ($\downarrow$) as a function of Chipformer online iterations. SGF is trained offline only.}
% %\label{table:wirelength_results}
% % \begin{tabular}{l|c|c}
% \toprule
% \textbf{Model} & \textbf{\textit{ami33}} & \textbf{\textit{ami49}} \\
% \midrule
% Chipformer   & 222,342 & 39,442 \\
% Chipformer (+5 iters) & 237,141 & 39,035 \\ %37184
% Chipformer (+10 iters) & 237,150 & 39,899 \\
% Chipformer (+15 iters) & 221,041 & 40,414 \\ % we added this after the submission deadline
% SGF (offline only)    & \textbf{200,738} & \textbf{28,636} \\
% \bottomrule
% \end{tabular}
% %\end{table}%

\begin{table}[th]
\centering
\caption{Manhattan wirelength ($\downarrow$) across benchmarks. }
\label{table:wirelength_results}
\begin{tabular}{lcccc}
\toprule
\textbf{Model} & \textbf{\textit{ami49}} & \textbf{\textit{n50}} & \textbf{\textit{n100}} & \textbf{\textit{ibm6}} \\
\midrule
Chipformer                   & 37{,}530 & 13{,}447 & 25{,}229 & 17{,}802 \\
SGF            & \textbf{32{,}304} & \textbf{8{,}747} & \textbf{16{,}607} & \textbf{16{,}321} \\
\bottomrule
\end{tabular}
\end{table}

Table \ref{table:wirelength_results_ami33} demonstrates the importance of spatial generalization. Prior art does not exploit the $L$-action similarity of the floorplanning problem. As shown, Chipformer’s wirelength fluctuates despite extended online fine-tuning, indicating that discrete action representations struggle to achieve consistent improvement through additional rollouts. In contrast, SGF achieves a lower wirelength after offline-only training, suggesting that continuous action representations can capture the underlying spatial structure more efficiently. Another advantage of SGF is that it decouples model output dimensionality from the resolution of the floorplanning canvas. By operating in a continuous coordinate space and deferring discretization to a $k$-NN step, SGF reduces the combinatorial blow-up afflicting the floorplanning.

\begin{table}[th]
\centering
\caption{Manhattan wirelength ($\downarrow$) on \textit{ami33} after permitting Chipformer to collect additional online rollouts for fine-tuning.}
\label{table:wirelength_results_ami33}
\begin{tabular}{lc}
\toprule
\textbf{Model} & \textbf{\textit{ami33}} \\
\midrule
SGF            & \textbf{202{,}617} \\
Chipformer                   & 221{,}539 \\
Chipformer (+50 rollouts)        & 221{,}670 \\
Chipformer (+100 rollouts)       & 224{,}539 \\
Chipformer (+150 rollouts)       & 220{,}051 \\
Chipformer (+200 rollouts)       & 221{,}539 \\
Chipformer (+250 rollouts)       & 224{,}546 \\
\bottomrule
\end{tabular}
\end{table}

As shown in Figure~\ref{fig:critic_test}, the error of critic predictions \textit{tends} to decrease as more modules are placed--indicating increasing certainty about final outcomes. However, we observe a small rise in wirelength prediction error towards the placement of the last few modules (placement of modules 21-33). Empirically, this corresponds to a phase where unplaced modules are more likely to be situated near the boundaries of the canvas. Since placements near the edges can significantly affect routing paths, the final wirelength becomes harder to predict—introducing more uncertainty. This behavior mirrors human layout planning, where early and middle stages benefit from clearer spatial context, while later-stage decisions are more ambiguous. Our experiments point to several broader takeaways:

\begin{itemize}
  \item \textbf{Representation matters.} In continuous output spaces, the model receives gradient signals across a region of similar actions, rather than a single discrete class. This inductive bias enables generalization from poor trajectories and smoother reward learning.

    \item \textbf{Structure can emerge from inductive bias.} SGF learns the consequences of spatial decisions through reward feedback.

  \item \textbf{Output-space decoupling reduces scaling burdens.} SGF avoids canvas-size-dependent output layers, enabling tractable reasoning in large 3D placement spaces.
  
  \item \textbf{Critic models can guide generation.} As we show in eq. \ref{eq:critic_hurts_q}, the return difference is bounded by the error of the critic. Having a model that can predict the performance of a floorplan can be exploited to generate one.
\end{itemize}

We would like to underscore that the primary contribution of this work is the foundational observation regarding floorplanning being an instance of an SLDAS. While we evaluated SGF on the widely used floorplanning benchmarks, we recognize that broader experimentation would strengthen empirical claims. Limitations are discussed further in the Appendix.

\section{Appendix}\label{sec: appendix}

\subsection{Model and Experiment Parameters}

\begin{table}[b]
\centering
\caption{Actor/Critic Input Encoders}
\setlength{\tabcolsep}{4pt}
\begin{tabular}{l|p{0.64\columnwidth}}
\toprule
\textbf{Input} & \textbf{Encoding} \\
\midrule
Return-to-Go & Linear $\rightarrow$ temporal self-attention \\
Grid & 3D CNN $\rightarrow$ Flatten $\rightarrow$ Linear $\rightarrow$ Tanh \\
Graph & GATConv $\rightarrow$ readout/pooling \\
Module Features & Linear $\rightarrow$ Tanh \\
Difference Map & 3D CNN encoder (same as Grid) \\
Actions & Linear $\rightarrow$ Tanh \\
Timesteps & Learned embedding \\
\bottomrule
\end{tabular}

\vspace{2pt}
{\raggedright\footnotesize We define $d_{\text{embed}}=d_{\text{model}}/6$. We use multi-head attention to encode the RTG to model interactions among wirelength, thermals, and congestion.\par}
\label{table:input_encoding}
\end{table}

Chipformer was trained offline for 200 epochs on each benchmark. It was subsequently fine-tuned online using its warm-up buffer $\Gamma_{\text{online}}$, over 5 iterations of 10 rollouts each, with 10 epochs of fine-tuning per rollout. Both training phases used the AdamW optimizer with learning rate $\eta = 6 \times 10^{-4}$ \cite{adamW}. The entropy weight parameters $(\beta, \kappa)$ were set to (0.5, 0.5) for \textit{ami33} and adjusted to (0.7, 0.7) for \textit{ami49} based on validation performance. 

SGF was trained for 200 offline epochs using the AdamW optimizer with learning rates of $5 \times 10^{-4}$ for the actor and $5 \times 10^{-3}$ for the critic. The context length was set to $n_{\text{modules}} + 1$, and inference results were computed as the best of three sampled trajectories for each model and netlist. Our canvas shape was $48\times 48 \times 3$ and we used $k=5$ for $k$-NN.

\subsection{Limitations}
Our goal is to study how continuous action representations can improve 3D floorplanning via generalizable spatial reasoning, not to deliver production-ready layouts. We use a fixed canvas and rectangular modules, simplifying geometry relative to commercial flows \cite{fp_harmful}. Power, frequency, interface constraints, and detailed physical effects are not explicitly modeled. Overall, we present a framework for how structure can emerge from inductive biases in learning systems; future work will incorporate richer constraints and benchmarks.

\subsection{Adapting Chipformer for 3D Placement and RL Env.}
We follow Chipformer unless noted. Changes: (1) replace 2D grid encoder with 3D Conv; (2) use $d_{model}{=}140$ and context length $n_{\text{modules}}{+}1$; (3) swap VGAE for GAE\footnote{We found that this improved their performance.}; (4) treat each SAR as one timestep by concatenating circuit, grid, action, and a null RTG into a single $d_{model}$ token. Chipformer does not use RTGs.

We use lightweight, course proxies for congestion and thermals. Wirelength is computed with a 3D Manhattan metric:
\[
\text{dist}(a,b)=\mathcal{M}(a,b)\big(|a_x{-}b_x|+|a_y{-}b_y|+10|a_z{-}b_z|\big).
\]
where $\mathcal{M}$ is  the net count between modules. For inference, each benchmark is run three times and the best rollout reported. Following \cite{chipformer, mirhoseini}, the environment also provides the next module/orientation and a legality mask over the canvas.

\bibliographystyle{plain}
\bibliography{main.bib}
\vspace{12pt}

\end{document}